\documentclass[letterpaper, 10 pt, conference]{ieeeconf}  %
\usepackage{url}
\usepackage{cite}
\usepackage{balance}
\usepackage{graphicx}
\usepackage{amsfonts}
\usepackage{fancyhdr}
\usepackage{comment}
\usepackage{times}
\usepackage{amsmath}
\usepackage{changepage}
\usepackage{amssymb}
\usepackage{enumerate}
\usepackage{algorithm}
\usepackage{algorithmic}
\usepackage{bm}
\usepackage{subfigure}
\usepackage{calligra}
\usepackage{multirow}
\usepackage{booktabs}
\usepackage{booktabs}

\IEEEoverridecommandlockouts                              %
\title{\LARGE \bf
	A Novel Dual-Lidar Calibration Algorithm Using Planar Surfaces
}

\author{Jianhao Jiao$^{1*}$, Qinghai Liao$^{1*}$, Yilong Zhu$^{1}$, Tianyu Liu$^{2}$, Yang Yu$^{1}$,\\ Rui Fan$^{1}$, Lujia Wang$^{3}$, Ming Liu$^{1}$
\thanks{This work was supported by National Natural Science Foundation of China No. U1713211 and 61603376, the Shenzhen Science, Technology and Innovation Commission(SZSTI) JCYJ20170818153518789, the Research Grant Council of Hong Kong SAR Government, China, under Project No. 11210017, and No. 21202816, and was also supported by the Guangdong Innovation and Technology Fund No. 2018B050502009, awarded to Prof. Ming Liu and Dr. Lujia Wang.}
\thanks{$^{1}$J. Jiao, Q. Liao, Y. Zhu, Y. Yu, R. Fan, and M. Liu are with the Robotics and Multi-Perception Laborotary, Robotics Institute, The Hong Kong University of Science and Technology, Hong Kong SAR, China {\tt\small \{jjiao, qinghai.liao, Yzhubr, yyubj, eeruifan, eelium\}}@ust.hk. }
	\thanks{$^{2}$T. Liu is with Unity-Drive technology Inc, Shenzhen, China {\tt\small \{liutianyu\}}@unity-drive.com.}
	\thanks{$^{3}$L. Wang is with the Shenzhen Institutes of Advanced Technology, Chinese Academy of Sciences, Shenzhen, China {\tt\small \{lj.wang1\}}@siat.ac.cn.}
	\thanks{$^*$Equal contribution.}
	}

\def\hlinew#1{%
	\noalign{\ifnum0=`}\fi\hrule \@height #1 \futurelet
	\reserved@a\@xhline}

\begin{document}
\IEEEpeerreviewmaketitle
\maketitle
\thispagestyle{empty}
\pagestyle{empty}

\begin{abstract}
Multiple lidars are prevalently used on mobile vehicles for rendering a broad view to enhance the performance of localization and perception systems. 
However, precise calibration of multiple lidars is challenging since the feature correspondences in scan points cannot always provide enough constraints.
To address this problem, the existing methods require fixed calibration targets in scenes or rely exclusively on additional sensors. In this paper, we present a novel method that enables automatic lidar calibration without these restrictions. 
Three linearly independent planar surfaces appearing in surroundings is utilized to find correspondences.
Two components are developed to ensure the extrinsic parameters to be found: a closed-form solver for initialization and an optimizer for refinement by minimizing a nonlinear cost function. Simulation and experimental results demonstrate the high accuracy of our calibration approach with the rotation and translation errors smaller than $\textbf{0.05}\textbf{rad}$ and $\textbf{0.1}\textbf{m}$ respectively.
\end{abstract}

\section{Introduction}
\label{sec.introduction}

Accurate extrinsic calibration has gained importance for vehicles which are equipped with a large number of sensors.
Traditional manual calibration techniques, which require known calibration targets \cite{geiger2012automatic,xie2015online,zhou2018automatic,liao2017extrinsic,liao2018extrinsic}, suffer limited flexibility and tend not to scale well to multi-sensor configurations. 

The automatic calibration methods of 3D lidars will be focused on in this paper. With the development of mobile robots, lidars have become one of the most popular sensors for perceiving the environment. Thanks to their accuracy and stability in measuring distance, they are used in many applications \cite{shan2018lego, zhou2018voxelnet, yun2019fl3d}. 
However, much recent work prefers the configuration with multiple lidars rather than a single lidar because it can render a richer view of environments and offer denser measurements. Several problems such as occlusion and sparsity can be avoided. On any mobile platform containing multiple lidars, it is of paramount importance that sensors can be calibrated automatically. When the calibration is finished, all measurements will be correctly projected into a unified coordinate system. 

\begin{figure}
	\label{fig.platform_and_system} 
	\centering
	\subfigure[]
	{\label{fig.platform}\centering\includegraphics[width=0.41\textwidth]{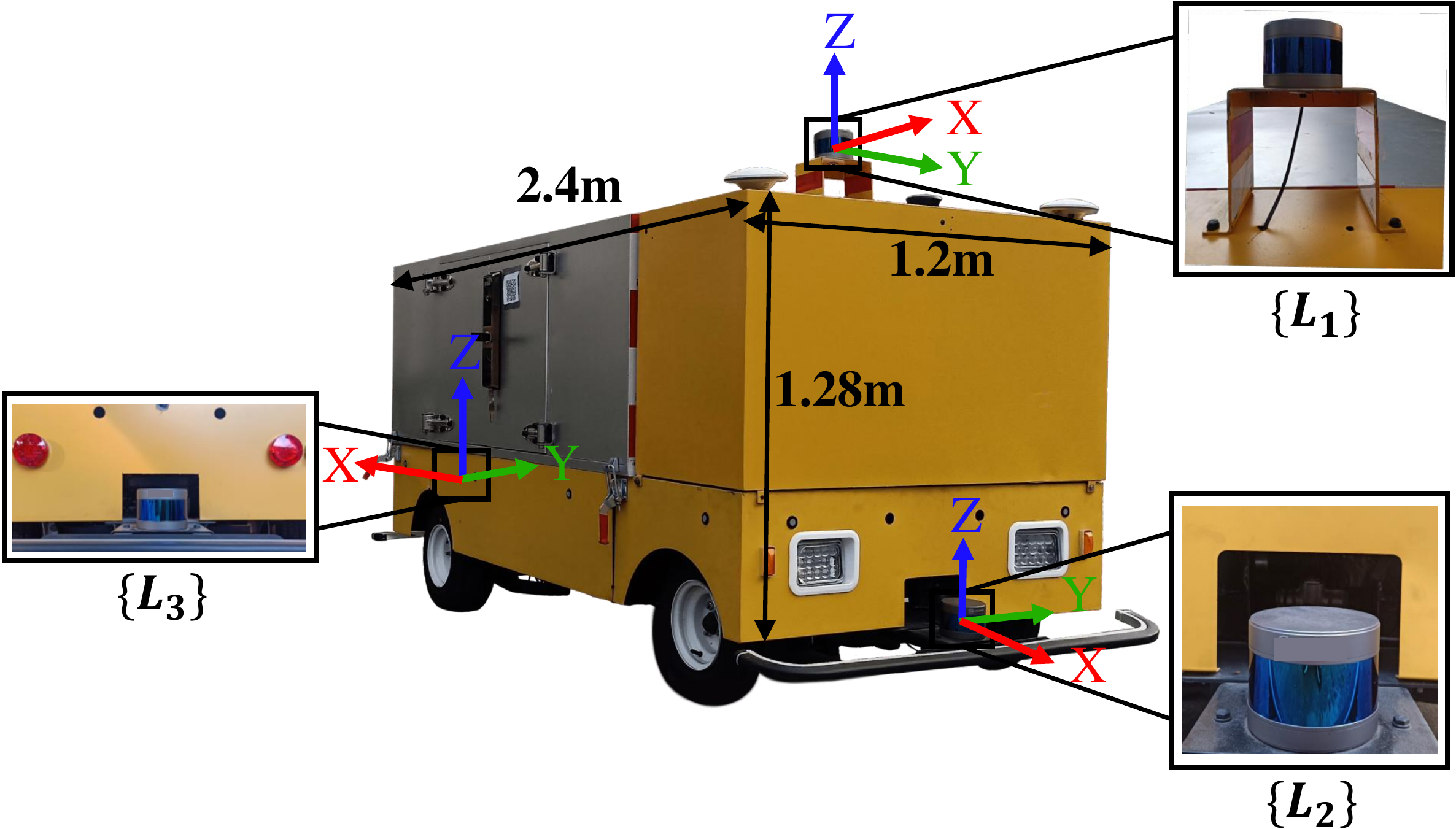}}    
	\subfigure[]
	{\label{fig.overview}\centering\includegraphics[width=0.42\textwidth]{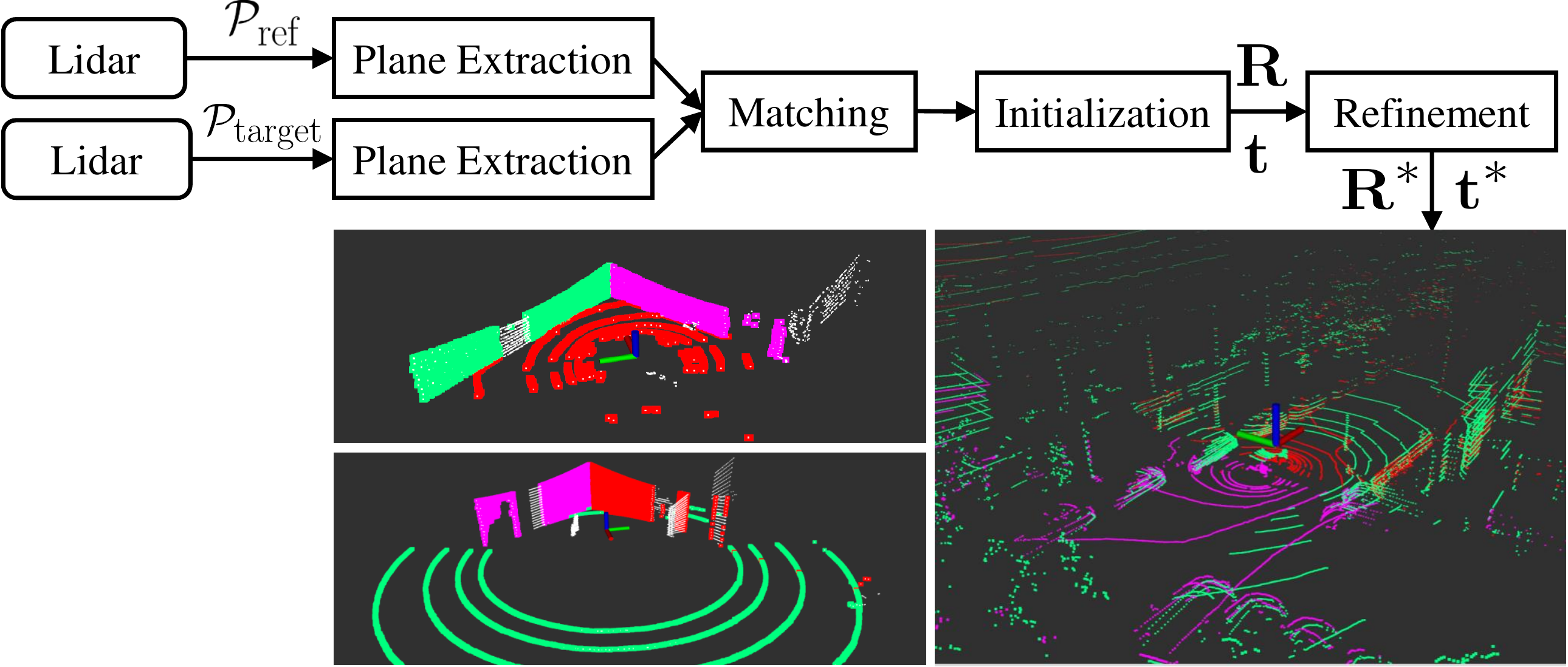}}    
	\caption{(a) Our mobile platform. (b) An overview of the proposed method. The bottom-left figure visualizes the extracted planes. The bottom-right figure visualizes the calibration results, where the red, pink and pink dots represent the points perceived by different lidars.}    
	\vspace{-0.5cm}
\end{figure}  

Over the past years, automatic methods for calibrating sensors e.g., lidar to camera \cite{taylor2013automatic}, multi-camera \cite{heng2013camodocal}, and camera to IMU \cite{yang2016self} have been proposed. However, few efforts have investigated multi-lidar calibration since this problem is challenging; as Choi et al. \cite{choi2016extrinsic} explained, ``searching for correspondences among scan points is difficult".
\cite{gao2010line} and \cite{he2013pairwise} are two specific multi-lidar calibration methods, but several drawbacks are presented.
Firstly, they rely exclusively on an additional sensor.
Secondly, their success will depend on the quality of initialization provided by users. 
Thirdly, both of them assume that mobile platforms should undergo efficient motion.

To tackle these issues, we propose a novel approach for calibrating dual lidars without any additional sensors and artificial markers.
This method assumes that three linearly independent planar surfaces forming a wall corner shape are provided as the calibration targets.
Through matching these planes, our method can successfully acquire the unknown extrinsic parameters in two steps: a closed-form solution for initialization and an optimizer for refinement by minimizing a defined cost function. This method is used to calibrate three lidars with overlapping regions on our mobile platform [see Fig. \ref{fig.platform}]. An overview of the method is shown in Fig. \ref{fig.overview}. In solving calibration with poor human intervention, we make two significant contributions in this paper:

\begin{itemize}
	\item We make it possible to use objects with unknown size in the outdoor environment as the calibration target.
	\item We demonstrate that our method is efficient in applications since the extrinsic parameters can be obtained immediately with one-shot measurement. 
\end{itemize}

The rest of the paper is organized as follows. In Section \ref{sec.related_work}, the related work is discussed. The methodology of our approach is introduced in Section \ref{sec.methodology}, followed by experimental results in Section \ref{sec.experiment}. Finally, Section \ref{sec.conclusion} summarizes the paper and discusses possible future work.

\section{Related Work}
\label{sec.related_work}

\subsection{Calibration for Multi-Lidar Systems}
In recent years, Gao and Spletzer \cite{gao2010line} proposed an algorithm to calibrate multiple lidars using point constraints provided by retro-reflective tapes in scenes. 
He et al. \cite{he2013pairwise} demonstrated a technique to extract geometric features among point clouds, which enables an offline algorithm to calibrate multiple 2D lidars in arbitrary scenes. Shortly after that, their approach was improved in a challenging scenario: an underground parking lot, where GPS is not available \cite{he2014calibration}. However, such methods rely on an additional localization module, making the calibration process complicated.

Artificial landmarks are prevalently used to find correspondences among sensor data. Xie et al. \cite{xie2018infrastructure} provided a general solution to jointly calibrate multiple cameras and lidars in the presence of a pre-built environment with apriltags. 
Steder et al. \cite{rowekamper2015automatic} proposed a tracking-based method to calibrate multiple 2D lidars using a moving object which appears in their overlapping areas. Based on it, Quenzel et al. \cite{quenzel2016robust} calibrated the same sensors with an additional verification step. However, these approaches require known markers to be placed in scenes.
In this paper, we exploit common planar surfaces as the calibration target inspired by \cite{choi2016extrinsic}, but our approach differs from it by releasing the orthogonal assumption of these surfaces to achieve outdoor calibration.

\subsection{Calibration for Other Sensing Systems}
There exist several published papers on lidar to camera, multi-camera and camera to IMU calibration. One of the first work to solve online camera and lidar calibration is \cite{levinson2013automatic}. 
In this method, edge features in images are associated with lidar measurements using depth discontinuities. The extrinsic parameters are optimized by minimizing a cost function. 
Different metrics based on Gradient Orientation Measure (GOM) \cite{taylor2013automatic}, Mutual Information (MI) \cite{pandey2015automatic}, and line-plane constraints \cite{zhou2018automatic} were also proposed.
However, all of them require initialization provided by users. In our proposed method, we introduce an algorithm to automatically initialize the extrinsic parameters by exploiting the geometric constraints of planar surfaces.

Developed from hand-eye calibration using the structure-from-motion techniques, motion-based approaches have been implemented to solve the calibration.
Heng et al. \cite{heng2013camodocal} proposed CamOdoCal, an automatic algorithm for four-camera calibration without the assumption of overlapping fields of view. They decouple the calibration process into initialization and refinement. In initialization, a rough estimate of extrinsic parameters is computed by combining visual odometry with the vehicle's egomotion. To refine the estimates, a bundle adjustment is used to optimize all of the cameras' poses and feature data. This pipeline is employed in our method. However, CamOdoCal was explicitly designed for vision sensors, which may not be feasible in various sensor configurations.
In contrast, Taylor and Nieto released a system \cite{taylor2015motion, taylor2016motion} to calibrate multiple heterogeneous sensors and their time offset. Generally, motion-based methods can work for a variety of configurations and can be integrated into several SLAM systems \cite{qin2018vins}. However, the calibration accuracy of motion-based methods is limited due to the drift of computed odometry, which needs to be refined using the appearance cues in the surroundings.

\section{Methodology}
\label{sec.methodology}

Our approach can make use of three linearly independent planar surfaces to calibration a dual-lidar system.
In this work, we introduce a robust algorithm to extract planes from scan points. The geometric structure of these planes can provide extrinsic parameters with enough constraints.
We define $\left\{L_{k}\right\}$ as a 3D coordinate system with its origin at the geometric center of a lidar $L_k$. The $x\textendash$, $y\textendash$ and $z\textendash$ axes are pointing forward, left and upward respectively. 
In this paper, we consider $\left\{L_{1}\right\}$ as the reference frame, and $\left\{L_k\right\}$ as the target frame. The point clouds perceived by a lidar is denoted by $\mathcal{P}$, and the coordinates of a point in $\mathcal{P}$ is represented as $\mathbf{p}_n = [x_n, y_n, z_n]^{\top}$.
Detailed notations are listed in Table \ref{tab.annotation} and are visualized in Fig. \ref{fig.lidar_plane}.

\begin{table}[b]
	\centering
	\caption{Annotation table.}
	\label{tab.annotation}    
	\begin{tabular}{cl}
		\hline 
		\toprule
		Notation         & Explanation                                           \\ \hline
		$\{L_1\}$ / $\{L_k\}$        & Reference / Target coordinate system                           \\ 
		$\Pi_i$ / $\Pi_{i}^{\prime}$         & $i^{th}$ planar surfaces in $\{L_1\}$ / $\{L_k\}$                  \\ 
		$\mathbf{o}$ / $\mathbf{o}^{\prime}$      & Intersection point in $\{L_1\}$ / $\{L_k\}$   \\ 
		$\bm{\beta}_{i}$ / $\bm{\beta}_{i}^{\prime}$ & Coefficients of $\Pi_i$ / $\Pi_{i}^{\prime}$ \\ 
		$\mathbf{n}_{i}$ / $\mathbf{n}_{i}^{\prime}$ & Unit normal vector of $\Pi_i$ / $\Pi_{i}^{\prime}$  \\
		\hline
		\toprule    \\  
	\end{tabular} 
\end{table}

\subsection{Plane Extraction}
\label{sec.plane_extraction}

\begin{figure}
	\centering    
	\includegraphics[width=0.4\textwidth]{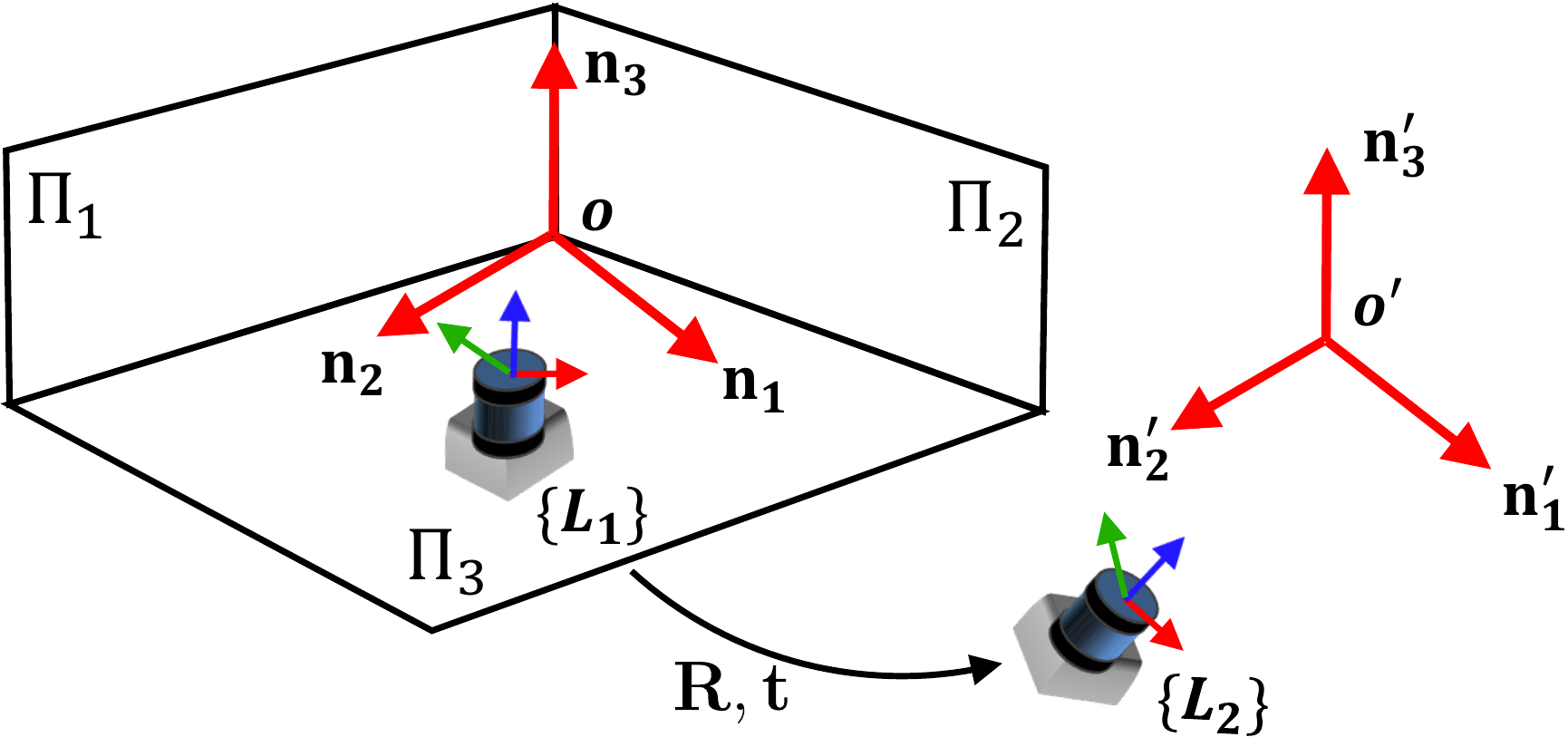} 
	\caption{A diagram of the notations. Red, green, and blue arrows denote the $x\textendash, y\textendash,$ and $z\textendash$ axes of each lidar coordinate system respectively.}
	\label{fig.lidar_plane}    
	\vspace{-0.3cm}                
\end{figure}    

\begin{figure}[]
	\centering    
	\includegraphics[width=0.4\textwidth]{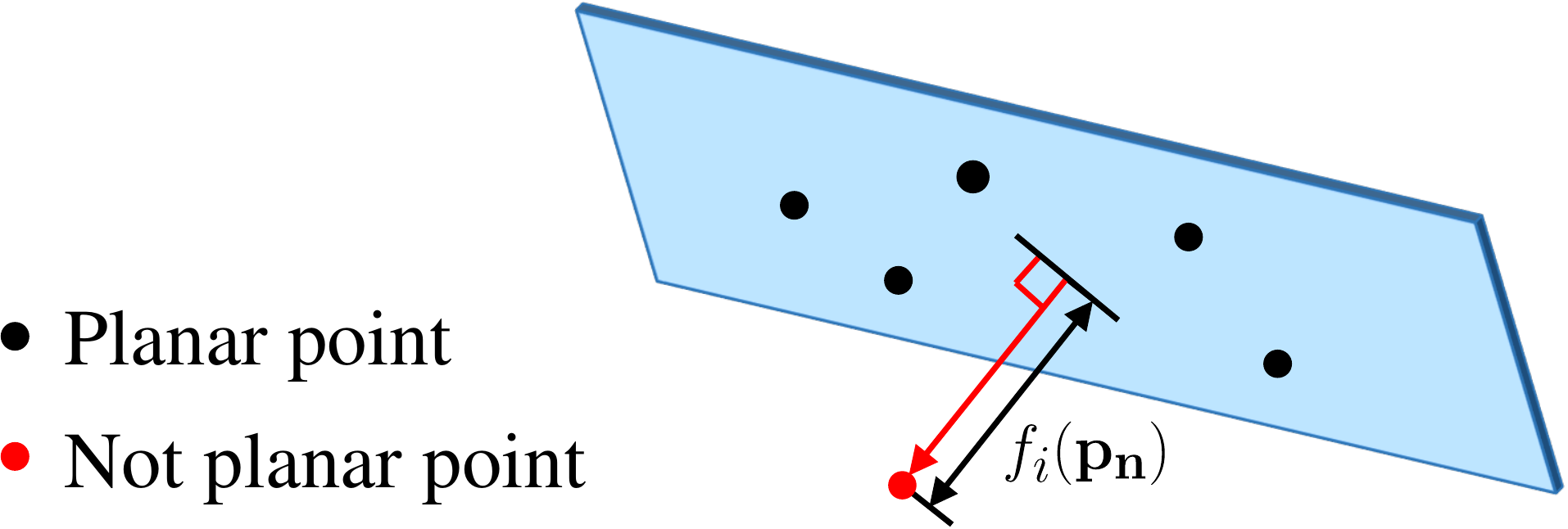} 
	\caption{Point to Plane distance.}
	\label{fig.point_plane_distance}          
	\vspace{-0.3cm}                              
\end{figure}

Denoting $\bm{\beta}_{i} = \left[\beta_{(i,0)},\beta_{(i,1)},\beta_{(i,2)},\beta_{(i,3)}\right]^{\top}$ the coefficients of $\Pi_i$, the distance between $\mathbf{p}_n$ and $\Pi_i$ [see Fig. \ref{fig.point_plane_distance}] is computed as follows:
\begin{equation}
\label{equ.plane_fitting}
\begin{aligned}
f_{i}(\mathbf{p}_n) = |\beta_{(i,0)}x_n+\beta_{(i,1)}y_n+\beta_{(i,2)}z_n+\beta_{(i,3)}|.
\end{aligned}    
\end{equation}

To fit a planar model from a series of discrete points, we employ the random sample consensus (RANSAC) algorithm.
By randomly selecting $N$ points from $\mathcal{P}$, the planar coefficients are acquired by solving a least-squares problem \cite{fan2019real}:
\begin{equation}
\label{equ.plane_fitting_2}
\begin{aligned}
\bm{\beta}_{i}^{*} = \underset{\bm{\beta}_{i}}{\arg \min}\sum_{n=1}^{N}f_{i}^{2}(\mathbf{p}_n),
\end{aligned}
\end{equation}
where the parameter vector $\bm{\beta}_{i}$ will be updated iteratively until an optimal model is acquired with maximum inlier points. To determine whether a point is an inlier, its square distance to a plane is computed. 
To extract three models, the RANSAC algorithm is executed separately at three times.
At each time, points belonging to former extracted models are ignored. Finally, we can obtain three groups of planar coefficients which are denoted by $\bm{\beta}_1$, $\bm{\beta}_2$, and $\bm{\beta}_3$ respectively to describe the planar surfaces. Hence, we can compute $\mathbf{o}=\left[o_x, o_y, o_z\right]^{\top}$ by solving a set of linear systems
\begin{equation}
\begin{aligned}
\begin{bmatrix}
\bm{\beta}_{1}^{\top} \\
\bm{\beta}_{2}^{\top} \\
\bm{\beta}_{3}^{\top} \\
\end{bmatrix}
\begin{bmatrix}
\mathbf{o} \\
1 \\
\end{bmatrix}
=
\textbf{0}.
\end{aligned}
\end{equation}

After computing $\bm{\beta}_{i}$, the unit normal vectors $\mathbf{n}_{1}$, $\mathbf{n}_2$ and $ \mathbf{n}_{3}$ can be represented up to scale. 
According to our assumption of linear independence, there exist three non-zero scalars $a$, $b$ and $c$ that satisfy the following equation:
\begin{equation}
\label{equ.normal_vector_direction}
a\mathbf{n}_1 + b\mathbf{n}_2 + c\mathbf{n}_3 = -\mathbf{o},
\end{equation}
where we can fix the directions of normal vectors to make $a$, $b$ and $c$ positive.

It is impossible to match these planar surfaces directly between lidars directly using the above results. Fig. \ref{fig.overview} shows an example of the extracted planes, where the color values represent their extraction order. We can observe that the corresponding planes do not have the same order.
By utilizing the wall corner shape [see Fig. \ref{fig.lidar_plane}], we find that these orders can be determined uniquely. Without loss of generality, we set $\Pi_1, \Pi_2$ as the \textit{left} and \textit{right} plane
respectively, and $\Pi_3$ as the \textit{bottom} plane. 
Their normal vectors should follow the right-hand rule:
\begin{equation} 
\label{equ.normal_order}
(\mathbf{n}_2 \times \mathbf{n}_1) \cdot \mathbf{n}_3 > 0.
\end{equation}

Following the above steps, we can correctly match the corresponding planes between two lidars.

\subsection{Initialization Using Closed-Form Solution}
\label{sec.initialization}
We can formulate the calibration of dual-lidar as a nonlinear optimization problem by minimizing the distance between corresponding planes. But the defined cost function is non-convex, as described in Section \ref{sec.optimization}. To avoid local minima, the parameters should be firstly initialized.

According to Section \ref{sec.plane_extraction}, we already have two sets of fitted planes $\Pi$ and $\Pi^{\prime}$ with known normal vectors. Consequently, their relative rotation $\mathbf{R}$ can be thus computed for initialization by introducing the Kabsch algorithm \cite{kabsch1978discussion}. The Kabsch algorithm is an effective approach that provides a least-squares solution to calculate the rotation between a pair of vector sets. We use $\mathbf{P}$, $\mathbf{Q}\in \mathbb{R}^{3}$ to indicate two $3\times 3$ matrices. Elements at the $i$th column of $\mathbf{P}$ are equal to $(\mathbf{n}_{i}-\mathbf{o})$, and these of $\mathbf{Q}$ is $(\mathbf{n}_{i}^{\prime}-\mathbf{o}^{\prime})$. We also denote $\mathbf{H}=\mathbf{P}^{\top}\mathbf{Q}$ the cross-covariance matrix. By calculating the singular value decomposition (SVD) of $\mathbf{H}=\mathbf{U}\mathbf{S}\mathbf{V}^{\top}$, $\mathbf{R}$ can be computed as:
\begin{equation}
\label{equ.initial_rotation}
\mathbf{R} = \mathbf{V}
\left(
\begin{matrix}
1 & 0 & 0 \\
0 & 1 & 0 \\
0 & 0 & d \\
\end{matrix}
\right)\mathbf{U}^{\top}, \ 
\text{where} \  d = \det(\mathbf{V}\mathbf{U}^{\top}).
\end{equation}

The relative translation $\mathbf{t}$ can be computed directly using the plane intersections:
\begin{equation}
\label{equ.initial_translation}
\mathbf{t} = \mathbf{o}^{\prime} - \mathbf{o}.
\end{equation}

\subsection{Nonlinear Optimization}
\label{sec.optimization}
The initial solution is further refined via a nonlinear optimization. By defining a cost function to describe the euclidean distance between $\Pi_i$ and $\Pi_{i}^{\prime}$, it can be computed as a sum of the squired distance  between a point $\mathbf{p}_{n}$ and its corresponding plane, i.e., $f_{i}^{2}(\mathbf{p}_n)$. We can write down the cost function and adopt a Levenberg-Marquardt algorithm for the nonlinear optimization:
\begin{equation}
\label{equ.refine_optimization}
\begin{split}
\mathbf{R}^{*}, \mathbf{t}^{*} 
&=
\underset{\mathbf{R},\mathbf{t}}{\arg \min}
\sum_{i=1}^{3}
\mathcal{F}(\mathbf{R}, \mathbf{t}, \Pi_{i}^{\prime}, \Pi_{i}^{}) \\
&= 
\underset{\mathbf{R},\mathbf{t}}{\arg \min}
\sum_{i=1}^{3}
\Bigg[
\sum_{\mathbf{p}^{\prime}\in\Pi_{i}^{\prime}}^{}
f_{i}^{2}\left(\mathbf{R}\mathbf{p}^{\prime}+\mathbf{t}\right) \\
& \ \ \ \ \ \ \ \ \ \ \ \ \ \ \ \ \ 
+
\sum_{\mathbf{p}\in\Pi_{i}}^{}
f_{i}^{\prime 2}\left(\mathbf{R}^{\prime}\mathbf{p}+\mathbf{t}^{\prime}\right)
\Bigg]^{2},
\end{split}
\end{equation}
where $f_{i}^{\prime}(\cdot)$ is the counterpart of $f_{i}$, and $\mathbf{R}^{\prime}=\mathbf{R}^{-1}$ as well as $\mathbf{t}^{\prime}=-\mathbf{R}^{-1}\mathbf{t}$ are a rotation matrix and a translation vector from $\left\{L_2\right\}$ to $\left\{L_1\right\}$ respectively.

\section{Experiment}
\label{sec.experiment}

To evaluate the proposed extrinsic calibration method, we test it with different configurations of dual lidars. Experiments are presented with synthetic data and real sensor data. All the resulting values are compared against the ground truth or the values provided by four methods in terms of accuracy.

\subsection{Implementation Details}
We adopt \verb|pcl|\footnote{\url{http://pointclouds.org}} to preprocess point clouds and implement the RANSAC-based plane fitting. \verb|Eigen|\footnote{\url{http://eigen.tuxfamily.org}} library is applied to implement the Kabsch algorithm, and \verb|Ceres Solver|\footnote{\url{http://ceres-solver.org}} is used to solve the nonlinear optimization problem. In the optimization, we set the maximum iteration as $1000$ and stopping tolerance as $1e^{-3}$.

\subsection{Experiments in Synthetic Data}
\label{sec.experiment_simulation}

\begin{figure}
	\label{fig.simulation_setup} 
	\centering
	\subfigure[Sythetic data.]
	{\label{fig.simulation_setup_1}\centering\includegraphics[width=0.22\textwidth]{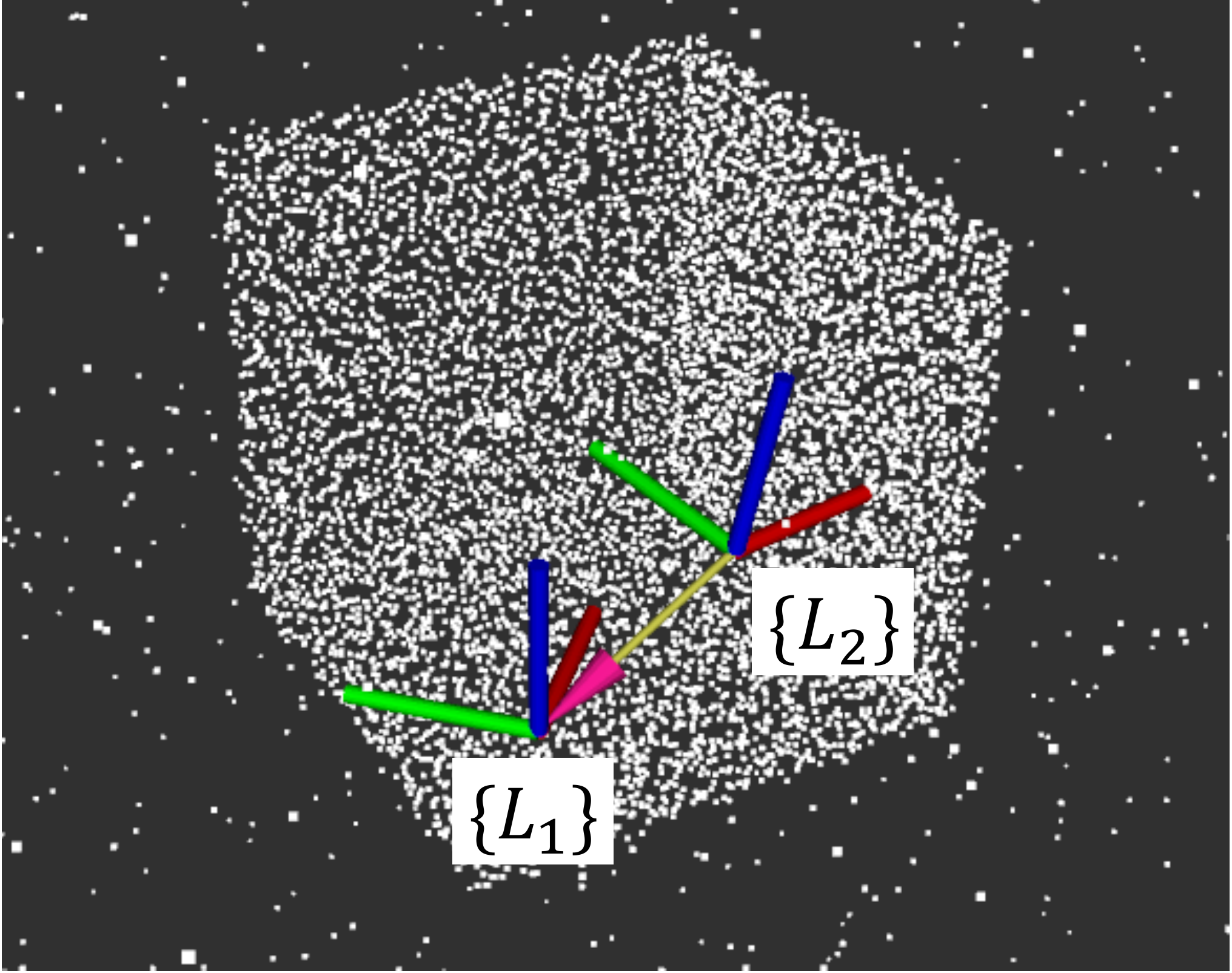}}
	\hspace{0.1cm}
	\subfigure[Calibrated point cloud.]
	{\label{fig.simulation_setup_2}\centering\includegraphics[width=0.20\textwidth]{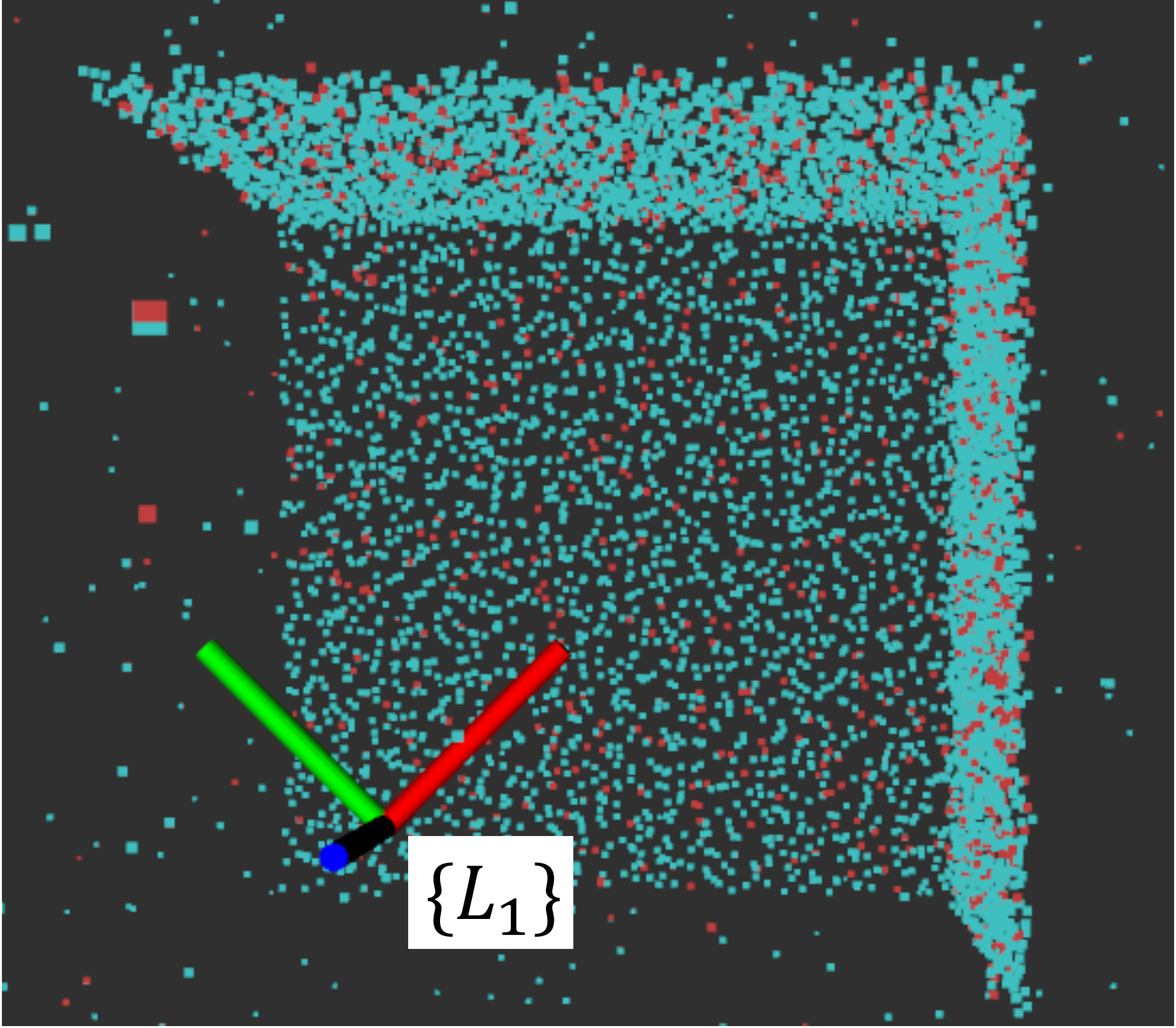}}       
	\caption{(a) An example of the synthetic data. (b) The red points and cyan points indicate the point cloud perceived by $L_1$ and $L_2$ respectively. We transform the cyan points from $\{L_2\}$ to $\{L_1\}$ using the results provided by our proposed method.}       
	\vspace{-0.3cm}                   
\end{figure}

To verify the performance of the proposed algorithm, we randomly generate $9500$ scan points ($7500$ planar points and $2000$ noisy points) in a $10m\times 10m\times 10m$ space. 
The planar points are generated evenly on three planar surfaces, which are subjected to zero-mean Gaussian noise with a standard deviation of $0.1m$. The rotation angles $\alpha$ on z-axis between $\Pi_1$ and $\Pi_2$ are set at intervals of $(60^{\circ}, 120^{\circ})$, while $\Pi_3$ is set on the bottom, which is orthogonal to $\Pi_1$ and $\Pi_2$. 
The noisy points are distributed in the space, which are subjected to zero-mean Gaussian distribution with a standard deviation of $5$m. $L_1$ is set arbitrarily where all the planar surfaces can be observed. Rotations from $\{L_1\}$ to $\{L_2\}$ are randomly generated within $(0^{\circ}, 20^{\circ})$, $(0^{\circ}, 20^{\circ})$, $(0^{\circ}, 360^{\circ})$ on $x\textendash$, $y\textendash$ and $z\textendash$ axis respectively, and translations are generated within $(-1.5, 1.5)m$ respectively. 
An example of the sensor configuration and the generated points is visualized in Fig. \ref{fig.simulation_setup_1}.
In our experiments, we randomly select two configurations with different $\mathbf{R}$ and $\mathbf{t}$ [see Table \ref{tab.simulation_result} (top)] as the ground truth to compare with the resulting values.

The difference in rotation is measured according to the angle difference between the ground truth $\mathbf{R}_{\text{gt}}$ and the resulting rotation $\mathbf{R}_{\text{res}}$, which is calculated as $e_{r} = \|\log(\mathbf{R}_{\text{gt}}^{}\mathbf{R}_{\text{res}}^{-1})^{\vee}\|_{2}$\footnote{The operator $\bm{\phi} = \log(\mathbf{R})^{\vee}$ is defined to associate $\mathbf{R}$ in $\text{SO}(3)$ to its rotation angle $\varphi\in\mathbb{R}^{3}$ on the axis.}. 
The difference in translation is computed using vector subtraction as $e_{t} = \|\mathbf{t}_{\text{gt}}^{}-\mathbf{t}_{\text{res}}^{}\|_{2}$.

\begin{figure}
	\centering
	\subfigure[Configuration 1]
	{\label{fig.simulation_result_group_1}\centering\includegraphics[width=0.46\textwidth]{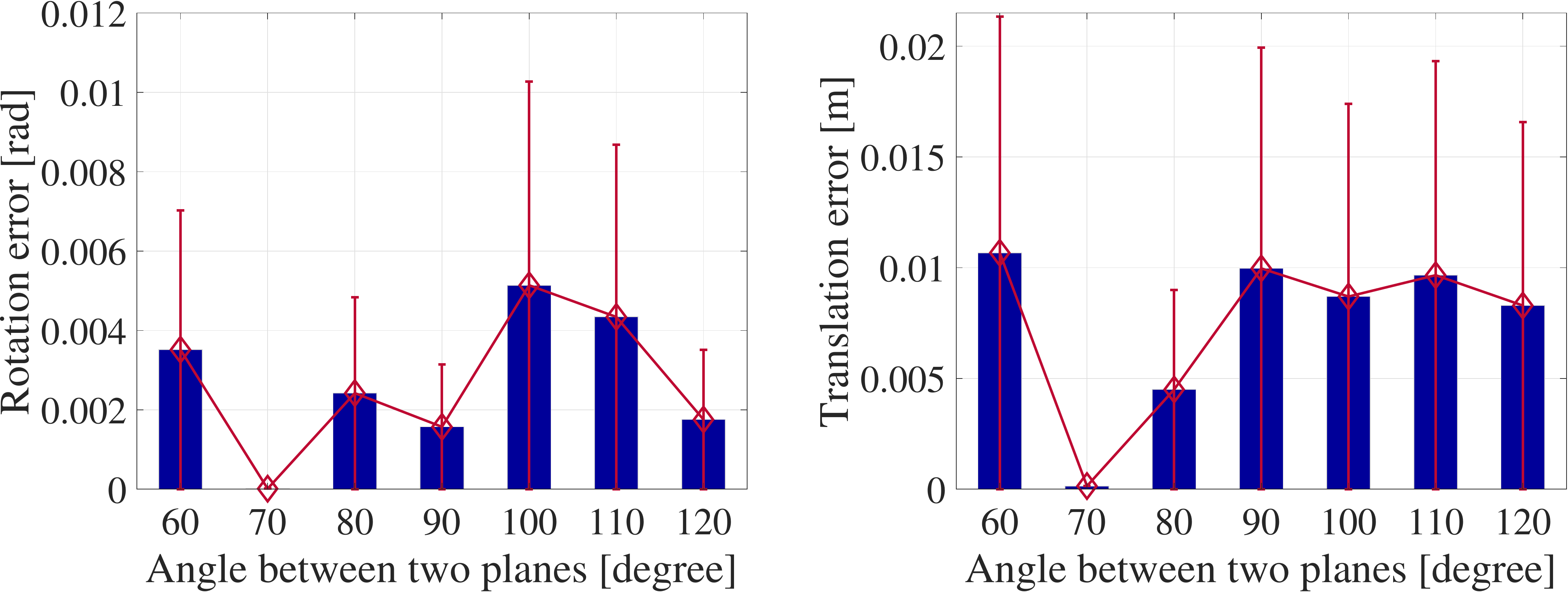}}
	\subfigure[Configuration 2]
	{\label{fig.simulation_result_group_2}\centering\includegraphics[width=0.46\textwidth]{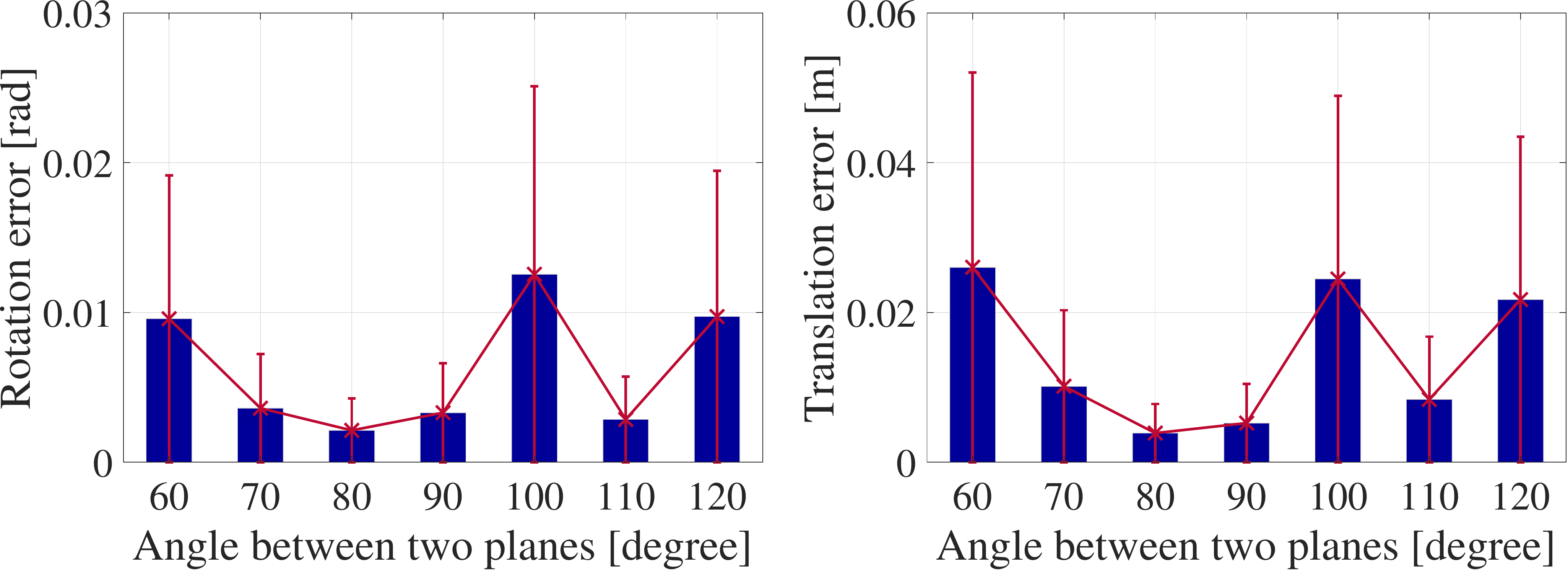}} 
	\caption{Performance analysis using rotation and translation errors on two sensor configurations. }                      
	\label{fig.simulation_result}         
	\vspace{-0.3cm}                       
\end{figure}

For each group of $\mathbf{R}$ and $\mathbf{t}$, we performed 10 trials on the noisy data and computed the mean as well as the standard deviation of the rotation and translation errors. In Fig. \ref{fig.simulation_result}, blue bars and red lines indicate the mean and standard deviation respectively. Detailed calibration results are shown in Table \ref{tab.simulation_result} (bottom). An example of the calibrated point cloud is shown in Fig. \ref{fig.simulation_setup_2}. In summary, we can see that the rotation and translation errors are tiny on the synthetic data. This proves that the proposed method can successfully calibrate the extrinsic parameters.

\begin{table}[]
	\centering
	\caption{The ground truth of two testing configurations (Top) and the calibration results (bottom).}
	\label{tab.simulation_result}    
	\begin{tabular}{ccc}
		\hline 
		\toprule
		\multicolumn{1}{c}{Conf.} & \multicolumn{1}{c}{Rotation {[$rad$]}} & \multicolumn{1}{c}{Translation {[$m$]}} \\ \hline
		\textbf{1}                           & 2.7337, -0.3946, -0.1809                & 0.8766, 0.4672, 1.0474                   \\
		\textbf{2}                           & -0.5174, 0.1277, 0.1222                 & 1.3785, -1.3929, 1.3020 \\          
		\hline
		\toprule    \\  
	\end{tabular}    
	
	\begin{tabular}{ccccccc}
		\hline 
		\toprule
		\multicolumn{1}{c}{\multirow{2}{*}{Conf.}}    & \multicolumn{2}{c}{\multirow{2}{*}{$\alpha$ [$degree$]}} & \multicolumn{2}{c}{Rotation Error [$rad$]}                                    & \multicolumn{2}{l}{Translation Error [$m$]}                                    \\ \cline{4-5} \cline{5-7}
		\multicolumn{1}{c}{}                          & \multicolumn{2}{c}{}                                                & \multicolumn{1}{c}{mean}   & \multicolumn{1}{c}{std.}    & \multicolumn{1}{c}{mean}   & \multicolumn{1}{c}{std.}    \\ \hline
		
		\multicolumn{1}{c}{\multirow{7}{*}{\textbf{1}}} & \multicolumn{2}{c}{60}                                             & \multicolumn{1}{c}{0.0035} & \multicolumn{1}{c}{0.0035} & \multicolumn{1}{c}{0.0107} & \multicolumn{1}{c}{0.0161} \\ 
		\multicolumn{1}{c}{}                        & \multicolumn{2}{c}{70}                                             & \multicolumn{1}{c}{0}      & \multicolumn{1}{c}{0}      & \multicolumn{1}{c}{0.0001} & \multicolumn{1}{c}{0}      \\ 
		\multicolumn{1}{c}{}                        & \multicolumn{2}{c}{80}                                             & \multicolumn{1}{c}{0.0024} & \multicolumn{1}{c}{0.0056} & \multicolumn{1}{c}{0.0045} & \multicolumn{1}{c}{0.0085} \\ 
		\multicolumn{1}{c}{}                        & \multicolumn{2}{c}{90}                                             & \multicolumn{1}{c}{0.0016} & \multicolumn{1}{c}{0.0040} & \multicolumn{1}{c}{0.0100} & \multicolumn{1}{c}{0.0243} \\ 
		\multicolumn{1}{c}{}                        & \multicolumn{2}{c}{100}                                            & \multicolumn{1}{c}{0.0051} & \multicolumn{1}{c}{0.0107} & \multicolumn{1}{c}{0.0087} & \multicolumn{1}{c}{0.0178} \\ 
		\multicolumn{1}{c}{}                        & \multicolumn{2}{c}{110}                                            & \multicolumn{1}{c}{0.0043} & \multicolumn{1}{c}{0.0063} & \multicolumn{1}{c}{0.0097} & \multicolumn{1}{c}{0.0161} \\ 
		\multicolumn{1}{c}{}                        & \multicolumn{2}{c}{120}                                            & \multicolumn{1}{c}{0.0018} & \multicolumn{1}{c}{0.0051} & \multicolumn{1}{c}{0.0083} & \multicolumn{1}{c}{0.0243} \\ \hline
		
		\multicolumn{1}{c}{\multirow{7}{*}{\textbf{2}}} & \multicolumn{2}{c}{60}                                             & \multicolumn{1}{c}{0.0096} & \multicolumn{1}{c}{0.0104} & \multicolumn{1}{c}{0.0260} & \multicolumn{1}{c}{0.0349} \\ 
		\multicolumn{1}{c}{}                        & \multicolumn{2}{c}{70}                                             & \multicolumn{1}{c}{0.0036} & \multicolumn{1}{c}{0.0083} & \multicolumn{1}{c}{0.0101} & \multicolumn{1}{c}{0.0274} \\ 
		\multicolumn{1}{c}{}                        & \multicolumn{2}{c}{80}                                             & \multicolumn{1}{c}{0.0021} & \multicolumn{1}{c}{0.0064} & \multicolumn{1}{c}{0.0039} & \multicolumn{1}{c}{0.0109} \\ 
		\multicolumn{1}{c}{}                        & \multicolumn{2}{c}{90}                                             & \multicolumn{1}{c}{0.0033} & \multicolumn{1}{c}{0.0071} & \multicolumn{1}{c}{0.0052} & \multicolumn{1}{c}{0.0101} \\ 
		\multicolumn{1}{c}{}                        & \multicolumn{2}{c}{100}                                            & \multicolumn{1}{c}{0.0126} & \multicolumn{1}{c}{0.0170} & \multicolumn{1}{c}{0.0245} & \multicolumn{1}{c}{0.0339} \\ 
		\multicolumn{1}{c}{}                        & \multicolumn{2}{c}{110}                                            & \multicolumn{1}{c}{0.0029} & \multicolumn{1}{c}{0.0057} & \multicolumn{1}{c}{0.0084} & \multicolumn{1}{c}{0.0163} \\ 
		\multicolumn{1}{c}{}                        & \multicolumn{2}{c}{120}                                            & \multicolumn{1}{c}{0.0097} & \multicolumn{1}{c}{0.0139} & \multicolumn{1}{c}{0.0217} & \multicolumn{1}{c}{0.0352} \\ 
		\hline \toprule
	\end{tabular}
\end{table}

\subsection{Experiments in Real Data}
\label{sec.experiment_read_data}

\begin{figure}
	\centering
	\subfigure[Buildings.]
	{\label{fig.easy_environment}\centering\includegraphics[width=0.214\textwidth]{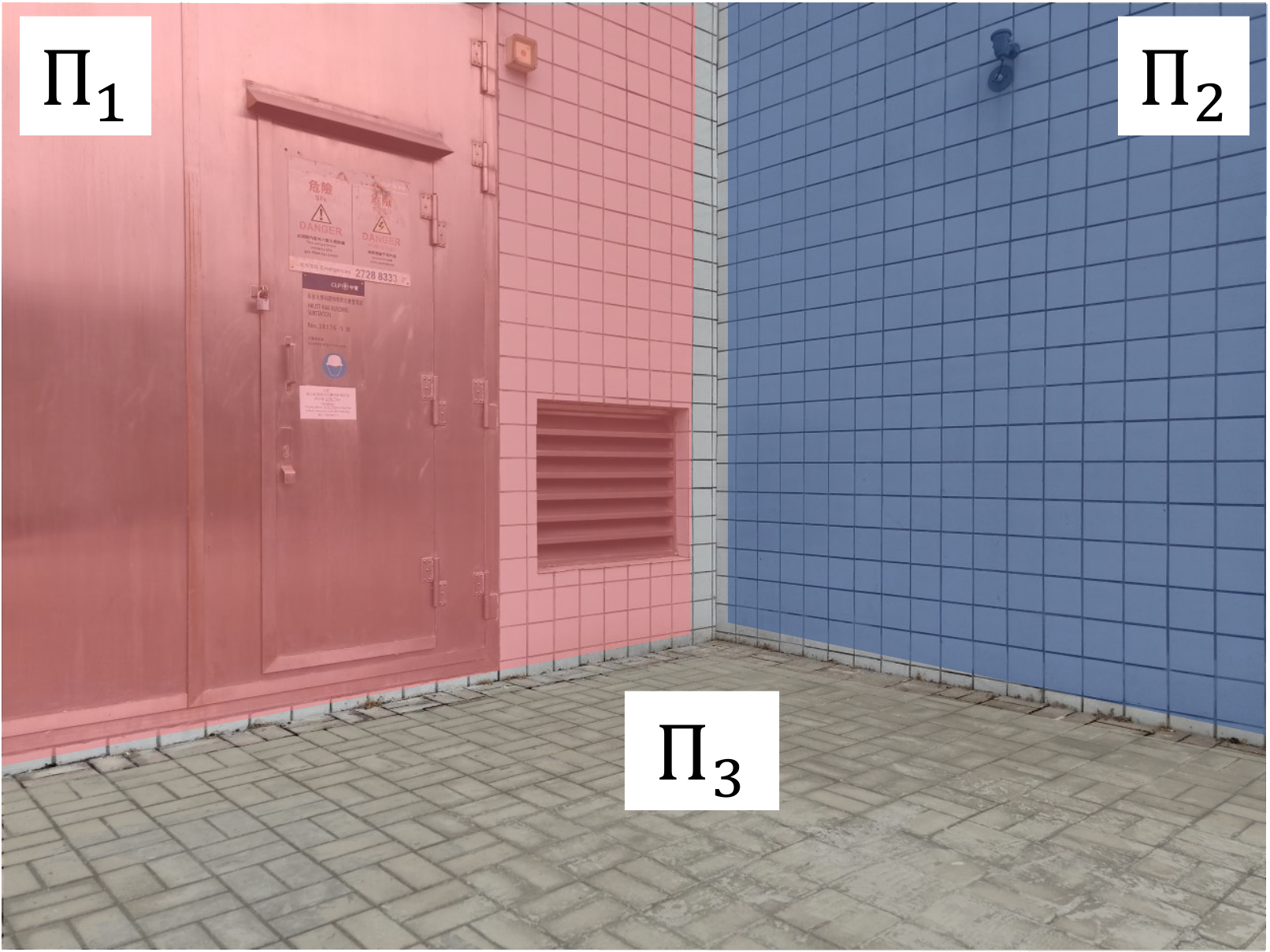}}
	\subfigure[Parking lot.]
	{\label{fig.challenging_environment}\centering\includegraphics[width=0.256\textwidth]{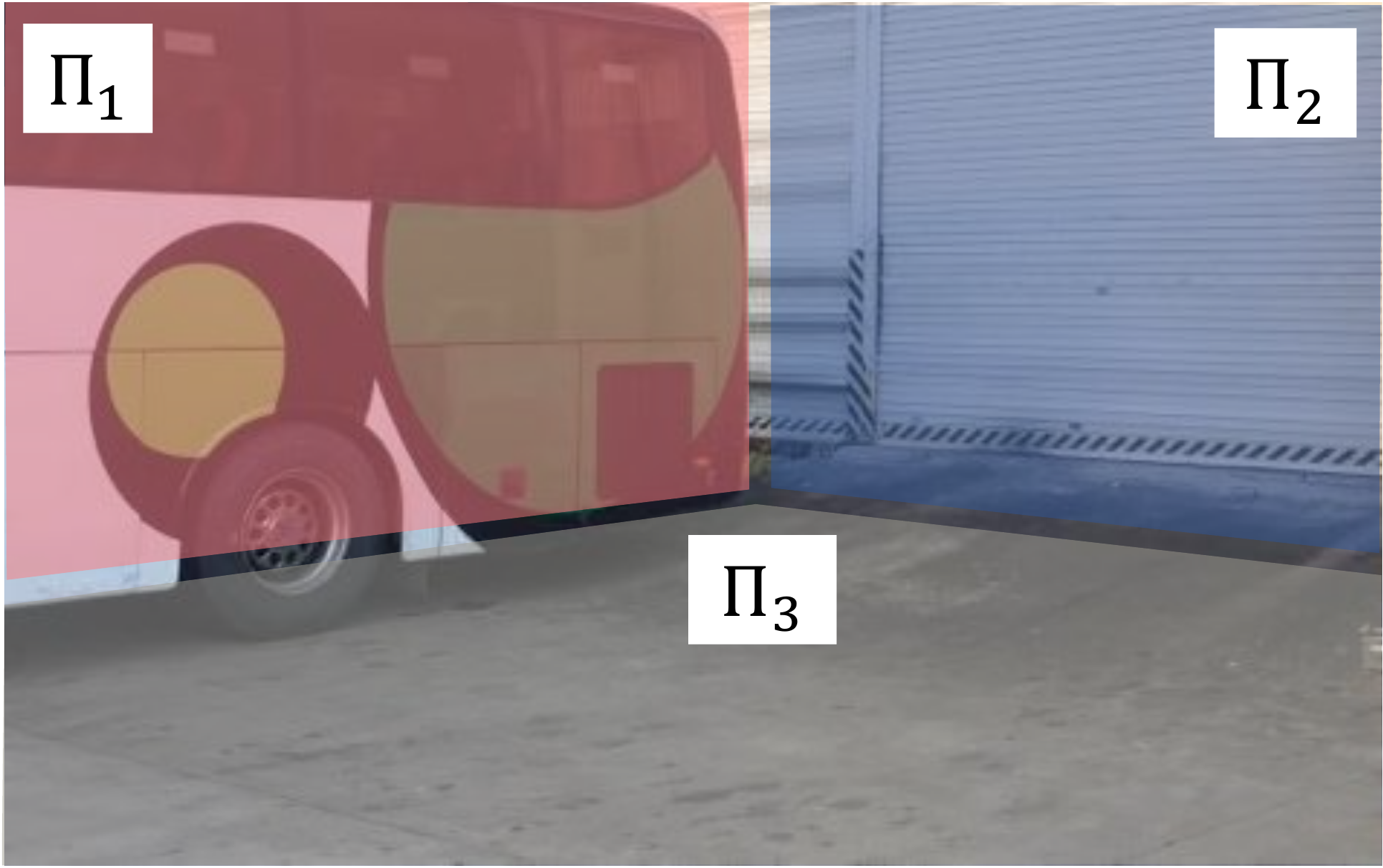}}
	\caption{The (a) \textit{easy} (b) \textit{hard} calibration environments which can be found in outdoors. They all form a wall corner shape. We extract three linear independent planar surfaces ($\Pi_1$, $\Pi_2$, and $\Pi_3$) for calibration.}  
	\label{fig.calibration_environment}  
\end{figure}  

\begin{figure}[]
	\centering    
	\includegraphics[width=0.48\textwidth]{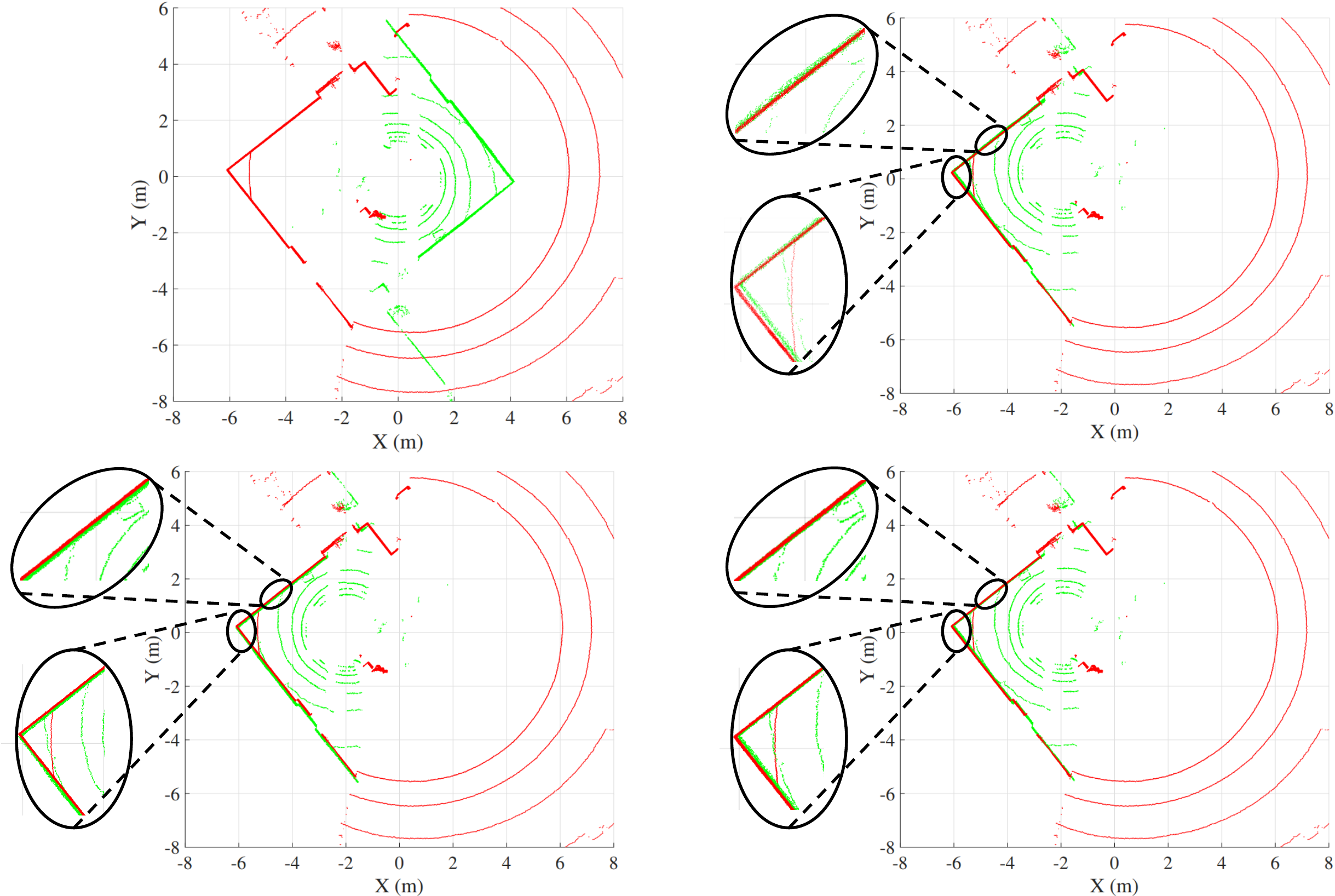}    
	\caption{Top views of the uncalibrated point clouds (top-left) and point clouds calibrated by W/O refinement (top-right), Ground truth (bottom-left) and Proposed (bottom-right) in the \textit{easy} scenario. The point cloud of $L_1$ and $L_3$ are denoted by red and green dots respectively.}
	\label{fig.calibrated_point_cloud}  
	\vspace{-0.3cm}                  
\end{figure}

We calibrate a sensor system which consists of three 16-beam RS-Lidars\footnote{\url{https://www.robosense.ai/rslidar/rs-lidar-16}} on our vehicle. As presented in Fig. \ref{fig.platform}, these lidars are mounted at the front ($L_1$), top ($L_2$), and tail ($L_3$) position respectively. Especially, $L_3$ is mounted with approximately $180^{\circ}$ rotation offset in yaw. In later sections, we denote $L_1 \ominus L_i$ the configuration between $L_1$ and $L_i$. The surrounding buildings and ground as the scan planar surfaces are used for calibration. 
We select two calibration environments in outdoor with two levels (\textit{easy} and \textit{hard}) for calibration, which are shown in Fig. \ref{fig.calibration_environment}. 

\subsubsection{\textit{Easy}}
Calibration is performed in the case of two different configurations: $L_1 \ominus L_2$, as a standard setup, and $L_1 \ominus L_3$, as a challenging setup. We take three methods for comparisons. The former two methods are developed based on the proposed one, but some steps are modified, while the last method is based on the motion-based techniques:
\begin{itemize}
	\item W/O refinement: The refinement step is removed.
	\item ICP refinement: The nonlinear optimization refinement is replaced by a point-to-plane ICP \cite{pomerleau2013comparing}.
	\item Motion-based: The motion of lidars are estimated using the LeGO-LOAM \cite{shan2018lego}, and the approach in \cite{taylor2016motion} is implemented to initialize the extrinsic parameters. In refinement, we use the approaches in \cite{pandey2017alignment}  based on ground surface to obtain the translation on $z\textendash$ axis.
\end{itemize}

Since the precise extrinsic parameters of the multi-LiDAR system are unknown, we use the values provided by the manufacturer as the ground truth to evaluate these methods.     

\begin{table}[]
	\caption{Estimated extrinsic parameters of $L_1 \ominus L_2$}
	\label{table.estimated_parameters_l1l2}    
	\centering
	\begin{tabular}{ccc}
		\hline \toprule
		\multicolumn{1}{c}{Method}         & \multicolumn{1}{c}{Rotation [$rad$]} & Error [$rad$]   \\ \hline
		\multicolumn{1}{c}{Ground truth} & \multicolumn{1}{c}{0.0096, 0.0989, 0.0425} & \multicolumn{1}{c}{---}\\
		\multicolumn{1}{c}{W/O refinement} & \multicolumn{1}{c}{-0.0046, -0.1138, -0.0193} & \multicolumn{1}{c}{0.2221}\\             
		\multicolumn{1}{c}{ICP refinement} & \multicolumn{1}{c}{-0.0882, -0.4254, 0.0462} & \multicolumn{1}{c}{0.5333}\\
		\multicolumn{1}{c}{Motion-based} & \multicolumn{1}{c}{0.2269, -0.0451, -0.0007} & \multicolumn{1}{c}{0.2332}\\        
		\multicolumn{1}{c}{Proposed}       & \multicolumn{1}{c}{0.0012, 0.0892, 0.0276} & \multicolumn{1}{c}{\textbf{0.0203}}\\ \hline \toprule
		\multicolumn{1}{c}{}               & \multicolumn{1}{c}{Translation [$m$]} & Error [$m$]  \\ \hline 
		Ground truth                     & \multicolumn{1}{c}{0.377002, -0.03309009, -1.23236}       & \multicolumn{1}{c}{---}\\
		W/O refinement                     & \multicolumn{1}{c}{0.314979, 0.0509327,  -1.23395}       & \multicolumn{1}{c}{0.102694}\\            
		ICP refinement                   & \multicolumn{1}{c}{1.21884, -0.0670357, -1.55964}               & \multicolumn{1}{c}{0.903941}      \\
		Motion-based                   & \multicolumn{1}{c}{0.2115, 0.4892, -1.1903}               & \multicolumn{1}{c}{0.5474}      \\        
		Proposed               & \multicolumn{1}{c}{0.336322, 0.00271319, -1.19076}   & \multicolumn{1}{c}{\textbf{0.067196}}\\ 
		\hline \toprule
	\end{tabular}
\end{table}

\begin{table}[]
	\caption{Estimated extrinsic parameters of $L_1 \ominus L_3$}
	\label{table.estimated_parameters_l1l3}    
	\centering
	\begin{tabular}{ccc}
		\hline \toprule
		\multicolumn{1}{c}{Method}         & \multicolumn{1}{c}{Rotation [$rad$]} & Error [$rad$] \\ \hline
		\multicolumn{1}{c}{Ground truth}  & \multicolumn{1}{c}{-0.0161, -0.0192, 3.1392} & \multicolumn{1}{c}{---} \\
		\multicolumn{1}{c}{W/O refinement} & \multicolumn{1}{c}{-0.0527, -0.0212, 3.1328} & \multicolumn{1}{c}{0.0373} \\
		\multicolumn{1}{c}{ICP refinement} & \multicolumn{1}{c}{-0.0333, -0.1329, 3.1277}  & \multicolumn{1}{c}{0.1605} \\
		\multicolumn{1}{c}{Motion-based} & \multicolumn{1}{c}{-0.0044, 0.0037, -2.6462}  & \multicolumn{1}{c}{0.4986} \\        
		\multicolumn{1}{c}{Proposed}       & \multicolumn{1}{c}{0.0004, -0.0020, 3.1375}  & \multicolumn{1}{c}{\textbf{0.0238}} \\ \hline \toprule
		\multicolumn{1}{c}{}               & \multicolumn{1}{c}{Translation [$m$]} &  Error [$m$]    \\ \hline
		Ground truth                     & \multicolumn{1}{c}{-1.96443, 0.04073154, -1.13756}       &  \multicolumn{1}{c}{---} \\
		W/O refinement                     & \multicolumn{1}{c}{-1.92115, 0.0968707,  -0.80199}          &  \multicolumn{1}{c}{0.341959} \\                
		ICP refinement                & \multicolumn{1}{c}{-1.95929, 0.0473068, -0.383619}          &  \multicolumn{1}{c}{0.753959} \\        
		Motion-based                & \multicolumn{1}{c}{-1.7794, -0.4670, -1.11136}          &  \multicolumn{1}{c}{0.5472} \\            
		Proposed               & \multicolumn{1}{c}{-1.9103, 0.052992, -1.0744}           &  \multicolumn{1}{c}{\textbf{0.08357}} \\ 
		\hline \toprule
	\end{tabular}
\end{table}

The results with respect to different configurations are listed in Table \ref{table.estimated_parameters_l1l2} and \ref{table.estimated_parameters_l1l3}. 
The estimated extrinsic parameters of the proposed algorithm are quite close to the ground truth. Regarding the relative rotation and translation errors, our method achieved [$0.0203rad, 0.067196m$] of $L_1 \ominus L_2$ and $[0.0238rad, 0.08357m]$ of $L_1 \ominus L_3$ respectively. We observe that larger errors are caused by ICP refinement and Motion-based methods. 
Regarding the ICP approach, failure is caused due to the wrong matching of corresponding planes. About the Motion-based approach, the drift and uncertainty of the estimated motion would significantly reduce the accuracy of the calibration results.
We can also see that the proposed method performs better than the w/o refinement method since the nonlinear optimization could further reduce the noise. During the calibration, the time for optimization is around $27.3s$ and $35.8s$ of the two configurations.

To evaluate the calibration results qualitatively, we transform the point cloud in $\{L_3\}$ to $\left\{L_1\right\}$ using the extrinsic parameters provided by ground truth, W/O refinement, and Proposed respectively. The top view of these fused point cloud can be seen in Fig. \ref{fig.calibrated_point_cloud}. We observe that the point clouds calibrated by the Proposed approach have litter uncertainty on the planar surfaces. Therefore, our algorithm can successfully calibrate dual lidars in real data with low error.

\subsubsection{\textit{Hard}}
\begin{figure}[]
	\centering    
	\includegraphics[width=0.47\textwidth]{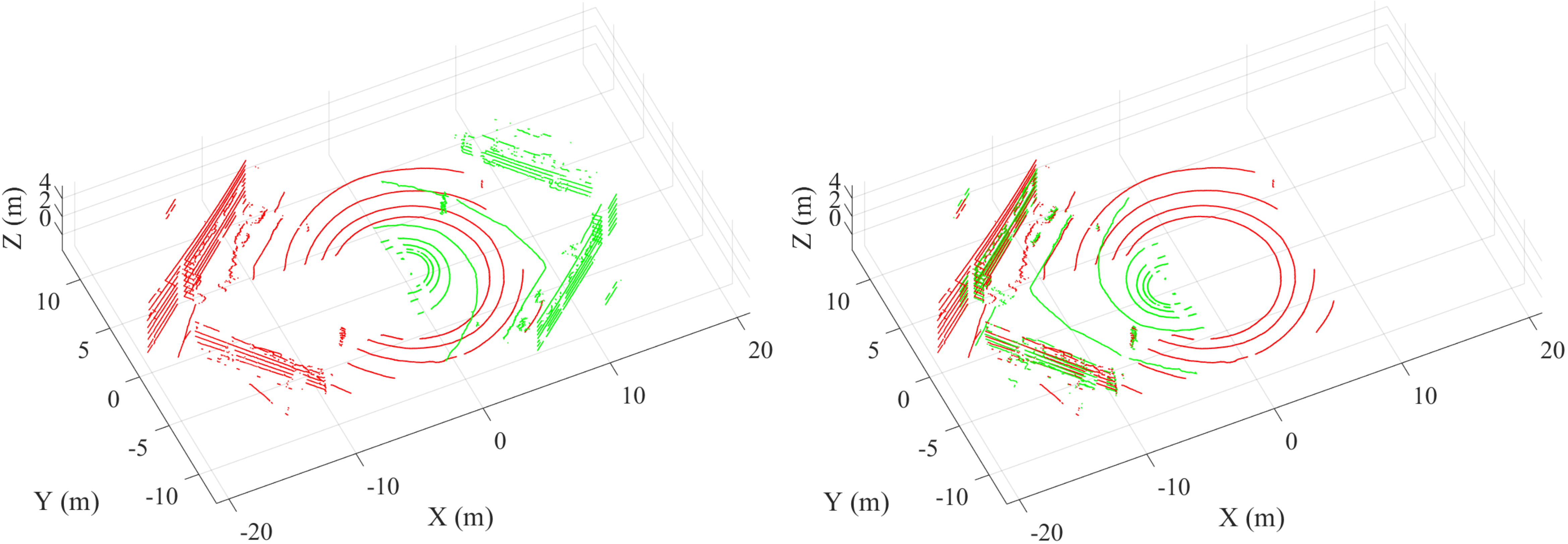}    
	\caption{Top views of the uncalibrated point clouds (left) and point clouds (right) calibrated by the proposed method in the \textit{hard} scenario. The point cloud of $L_1$ and $L_3$ are denoted by red and green dots respectively.. }
	\label{fig.challenging_calibration}  
	\vspace{-0.3cm}
\end{figure}

In the following, we study the performance of our approach in the \textit{hard} scenario. This scenario is more challenging because there are several objects on these planes that may influence the plane extraction and optimization results. In this experiment, the configuration $L_1 \ominus L_3$ is calibrated. 
The relative rotation and translation errors compared with the ground truth are [$0.0292rad$, $0.0930m$], while the optimization time is around $106.4s$. The fused point clouds are shown in Fig. \ref{fig.challenging_calibration}, where the planar surfaces are registered well without much offset. We conclude that the extrinsic parameters can be recovered in this scenario.

\subsection{Discussion}
\label{sec.discussion}

We have three certain assumptions in this method: lidars are horizontally mounted; they share large overlapping fields of view with each other and three planar surfaces are provided for calibration. 
Therefore, the proposed method may fail in several cases. For instance, if lidars are mounted at an arbitrary orientation, a wrong initialization will be caused. 
Another case is that if planar surfaces are wrongly detected, their correspondences will be mismatched.

\section{Conclusion and Future Work} 
\label{sec.conclusion}

In this paper, we have presented an automatic algorithm for calibrating a dual-lidar system without any additional sensors, artificial landmarks, or information about the motion provided by sensors. 
The RANSAC-based model fitting approach is used to extract three linearly independent planar surfaces from scan points. Linear constraints for initialization are provided by the geometric structure of these planar surfaces, and a final nonlinear optimization is then used to refine the estimates. 
Our proposed method has been demonstrated to recover the extrinsic parameters of a dual-lidar system with rotation and translation error smaller than $0.05rad$ and $0.1m$ in different testing conditions.

It would be beneficial to determine the parameters in more general cases, e.g., sensors do not share any overlapping fields of view, or they are arbitrarily mounted on a vehicle. Such problems may be solved by developing a simultaneous localization and mapping system, where the extrinsic parameters are jointly optimized with the odometry and map within a unified framework.

\balance
\bibliographystyle{IEEEtran}
\bibliography{reference}{}

\begin{thebibliography}{10}
\providecommand{\url}[1]{#1}
\csname url@samestyle\endcsname
\providecommand{\newblock}{\relax}
\providecommand{\bibinfo}[2]{#2}
\providecommand{\BIBentrySTDinterwordspacing}{\spaceskip=0pt\relax}
\providecommand{\BIBentryALTinterwordstretchfactor}{4}
\providecommand{\BIBentryALTinterwordspacing}{\spaceskip=\fontdimen2\font plus
\BIBentryALTinterwordstretchfactor\fontdimen3\font minus
  \fontdimen4\font\relax}
\providecommand{\BIBforeignlanguage}[2]{{%
\expandafter\ifx\csname l@#1\endcsname\relax
\typeout{** WARNING: IEEEtran.bst: No hyphenation pattern has been}%
\typeout{** loaded for the language `#1'. Using the pattern for}%
\typeout{** the default language instead.}%
\else
\language=\csname l@#1\endcsname
\fi
#2}}
\providecommand{\BIBdecl}{\relax}
\BIBdecl

\bibitem{geiger2012automatic}
A.~Geiger, F.~Moosmann, {\"O}.~Car, and B.~Schuster, ``Automatic camera and
  range sensor calibration using a single shot,'' in \emph{Robotics and
  Automation (ICRA), 2012 IEEE International Conference on}.\hskip 1em plus
  0.5em minus 0.4em\relax IEEE, 2012, pp. 3936--3943.

\bibitem{xie2015online}
G.~Xie, T.~Xu, C.~Isert, M.~Aeberhard, S.~Li, and M.~Liu, ``Online active
  calibration for a multi-lrf system,'' in \emph{2015 IEEE 18th International
  Conference on Intelligent Transportation Systems}.\hskip 1em plus 0.5em minus
  0.4em\relax IEEE, 2015, pp. 806--811.

\bibitem{zhou2018automatic}
L.~Zhou, Z.~Li, and M.~Kaess, ``Automatic extrinsic calibration of a camera and
  a 3d lidar using line and plane correspondences,'' in \emph{2018 IEEE/RSJ
  International Conference on Intelligent Robots and Systems (IROS)}.\hskip 1em
  plus 0.5em minus 0.4em\relax IEEE, 2018, pp. 5562--5569.

\bibitem{liao2017extrinsic}
Q.~Liao, M.~Liu, L.~Tai, and H.~Ye, ``Extrinsic calibration of 3d range finder
  and camera without auxiliary object or human intervention,'' 2017.

\bibitem{liao2018extrinsic}
Q.~Liao, Z.~Chen, Y.~Liu, Z.~Wang, and M.~Liu, ``Extrinsic calibration of lidar
  and camera with polygon,'' in \emph{2018 IEEE International Conference on
  Robotics and Biomimetics (ROBIO)}.\hskip 1em plus 0.5em minus 0.4em\relax
  IEEE, 2018, pp. 200--205.

\bibitem{shan2018lego}
T.~Shan and B.~Englot, ``Lego-loam: Lightweight and ground-optimized lidar
  odometry and mapping on variable terrain,'' 2018.

\bibitem{zhou2018voxelnet}
Y.~Zhou and O.~Tuzel, ``Voxelnet: End-to-end learning for point cloud based 3d
  object detection,'' in \emph{Proceedings of the IEEE Conference on Computer
  Vision and Pattern Recognition}, 2018, pp. 4490--4499.

\bibitem{yun2019fl3d}
P.~Yun, L.~Tai, Y.~Wang, C.~Liu, and M.~Liu, ``Focal loss in 3d object
  detection,'' \emph{IEEE Robotics and Automation Letters}, vol.~4, no.~2, pp.
  1263--1270, April 2019.

\bibitem{taylor2013automatic}
Z.~Taylor, J.~Nieto, and D.~Johnson, ``Automatic calibration of multi-modal
  sensor systems using a gradient orientation measure,'' in \emph{Intelligent
  Robots and Systems (IROS), 2013 IEEE/RSJ International Conference on}.\hskip
  1em plus 0.5em minus 0.4em\relax IEEE, 2013, pp. 1293--1300.

\bibitem{heng2013camodocal}
L.~Heng, B.~Li, and M.~Pollefeys, ``Camodocal: Automatic intrinsic and
  extrinsic calibration of a rig with multiple generic cameras and odometry,''
  in \emph{Intelligent Robots and Systems (IROS), 2013 IEEE/RSJ International
  Conference on}.\hskip 1em plus 0.5em minus 0.4em\relax IEEE, 2013, pp.
  1793--1800.

\bibitem{yang2016self}
Z.~Yang, T.~Liu, and S.~Shen, ``Self-calibrating multi-camera visual-inertial
  fusion for autonomous mavs,'' in \emph{Intelligent Robots and Systems (IROS),
  2016 IEEE/RSJ International Conference on}.\hskip 1em plus 0.5em minus
  0.4em\relax IEEE, 2016, pp. 4984--4991.

\bibitem{choi2016extrinsic}
D.-G. Choi, Y.~Bok, J.-S. Kim, and I.~S. Kweon, ``Extrinsic calibration of 2-d
  lidars using two orthogonal planes,'' \emph{IEEE Transactions on Robotics},
  vol.~32, no.~1, pp. 83--98, 2016.

\bibitem{gao2010line}
C.~Gao and J.~R. Spletzer, ``On-line calibration of multiple lidars on a mobile
  vehicle platform,'' in \emph{Robotics and Automation (ICRA), 2010 IEEE
  International Conference on}.\hskip 1em plus 0.5em minus 0.4em\relax IEEE,
  2010, pp. 279--284.

\bibitem{he2013pairwise}
M.~He, H.~Zhao, F.~Davoine, J.~Cui, and H.~Zha, ``Pairwise lidar calibration
  using multi-type 3d geometric features in natural scene,'' in
  \emph{Intelligent Robots and Systems (IROS), 2013 IEEE/RSJ International
  Conference on}.\hskip 1em plus 0.5em minus 0.4em\relax IEEE, 2013, pp.
  1828--1835.

\bibitem{he2014calibration}
M.~He, H.~Zhao, J.~Cui, and H.~Zha, ``Calibration method for multiple 2d lidars
  system,'' in \emph{Robotics and Automation (ICRA), 2014 IEEE International
  Conference on}.\hskip 1em plus 0.5em minus 0.4em\relax IEEE, 2014, pp.
  3034--3041.

\bibitem{xie2018infrastructure}
Y.~Xie, R.~Shao, P.~Guli, B.~Li, and L.~Wang, ``Infrastructure based
  calibration of a multi-camera and multi-lidar system using apriltags,'' in
  \emph{2018 IEEE Intelligent Vehicles Symposium (IV)}.\hskip 1em plus 0.5em
  minus 0.4em\relax IEEE, 2018, pp. 605--610.

\bibitem{rowekamper2015automatic}
J.~R{\"o}wek{\"a}mper, M.~Ruhnke, B.~Steder, W.~Burgard, and G.~D. Tipaldi,
  ``Automatic extrinsic calibration of multiple laser range sensors with little
  overlap,'' in \emph{Robotics and Automation (ICRA), 2015 IEEE International
  Conference on}.\hskip 1em plus 0.5em minus 0.4em\relax IEEE, 2015, pp.
  2072--2077.

\bibitem{quenzel2016robust}
J.~Quenzel, N.~Papenberg, and S.~Behnke, ``Robust extrinsic calibration of
  multiple stationary laser range finders,'' in \emph{Automation Science and
  Engineering (CASE), 2016 IEEE International Conference on}.\hskip 1em plus
  0.5em minus 0.4em\relax IEEE, 2016, pp. 1332--1339.

\bibitem{levinson2013automatic}
J.~Levinson and S.~Thrun, ``Automatic online calibration of cameras and
  lasers.'' in \emph{Robotics: Science and Systems}, vol.~2, 2013.

\bibitem{pandey2015automatic}
G.~Pandey, J.~R. McBride, S.~Savarese, and R.~M. Eustice, ``Automatic extrinsic
  calibration of vision and lidar by maximizing mutual information,''
  \emph{Journal of Field Robotics}, vol.~32, no.~5, pp. 696--722, 2015.

\bibitem{taylor2015motion}
Z.~Taylor and J.~Nieto, ``Motion-based calibration of multimodal sensor
  arrays,'' in \emph{Robotics and Automation (ICRA), 2015 IEEE International
  Conference on}.\hskip 1em plus 0.5em minus 0.4em\relax IEEE, 2015, pp.
  4843--4850.

\bibitem{taylor2016motion}
Z.~Taylor and J.~Niet, ``Motion-based calibration of multimodal sensor
  extrinsics and timing offset estimation,'' \emph{IEEE Transactions on
  Robotics}, vol.~32, no.~5, pp. 1215--1229, 2016.

\bibitem{qin2018vins}
T.~Qin, P.~Li, and S.~Shen, ``Vins-mono: A robust and versatile monocular
  visual-inertial state estimator,'' \emph{IEEE Transactions on Robotics},
  vol.~34, no.~4, pp. 1004--1020, 2018.

\bibitem{fan2019real}
R.~Fan, J.~Jiao, J.~Pan, H.~Huang, S.~Shen, and M.~Liu, ``Real-time dense
  stereo embedded in a uav for road inspection,'' 2019.

\bibitem{kabsch1978discussion}
W.~Kabsch, ``A discussion of the solution for the best rotation to relate two
  sets of vectors,'' \emph{Acta Crystallographica Section A: Crystal Physics,
  Diffraction, Theoretical and General Crystallography}, vol.~34, no.~5, pp.
  827--828, 1978.

\bibitem{pomerleau2013comparing}
F.~Pomerleau, F.~Colas, R.~Siegwart, and S.~Magnenat, ``Comparing icp variants
  on real-world data sets,'' \emph{Autonomous Robots}, vol.~34, no.~3, pp.
  133--148, 2013.

\bibitem{pandey2017alignment}
G.~Pandey, S.~Giri, and J.~R. Mcbride, ``Alignment of 3d point clouds with a
  dominant ground plane,'' in \emph{Intelligent Robots and Systems (IROS), 2017
  IEEE/RSJ International Conference on}.\hskip 1em plus 0.5em minus 0.4em\relax
  IEEE, 2017, pp. 2143--2150.

\end{thebibliography}

\end{document}